\documentclass[runningheads]{llncs}
\usepackage{graphicx}
\usepackage{amsfonts} 

\usepackage[noadjust]{cite}

\begin{document}
\title{Automatic Hand Sign Recognition:
\\ 
Identify Unusuality through Latent Cognizance}
%
%
\author{Pisit Nakjai\orcidID{0000-0001-8471-4390} 
\and
Tatpong Katanyukul\orcidID{0000-0003-3586-475X}}
\authorrunning{P. Nakjai and T. Katanyukul}
%
\institute{Khon Kaen University, Khon Kaen, Thailand
\\
\email{mynameisbee@gmail.com, tatpong@kku.ac.th}
}
%
\maketitle              
\begin{abstract}
Sign language is a main communication channel among hearing disability community.
Automatic sign language transcription could facilitate better communication and understanding between hearing disability community and hearing majority.

As a recent work in automatic sign language transcription has discussed, effectively handling or identifying a non-sign posture is one of the key issues.
A non-sign posture is a posture unintended for sign reading and does not belong to any valid sign.
A non-sign posture may arise during sign transition or simply from an unaware posture. 
Confidence ratio has been proposed to mitigate the 
issue. 
Confidence ratio is simple to compute and readily available without extra training.
However, confidence ratio is reported to only partially address the problem. In addition, confidence ratio formulation is susceptible to computational instability. 

This article proposes alternative formulations to confidence ratio, investigates an issue of non-sign identification for Thai Finger Spelling recognition,
 explores potential solutions 
 and has found a promising direction. 
Not only does this finding address the issue of non-sign identification, it also provide some insight behind a well-learned inference machine, revealing hidden meaning and new interpretation of the underlying mechanism. 
Our proposed methods are evaluated and shown to be effective 
for non-sign detection.

\end{abstract}

\section{Introduction}

Sign language is a main face-to-face communication channel in a hearing disability community.
Like spoken languages, there are many sign languages, e.g., American Sign Language (ASL), British Sign Language (BSL), French Sign Language (LSF), Spanish Sign Language (LSE), 
Italian Sign Language (LIS), 
German Sign Language (DGS),
Chinese Sign Language (CSL),
Japanese Sign Language (JSL),
Indo-Pakistani Sign Language (IPSL),
Thai Sign Language (TSL), etc.
A sign language usually has two schemes: semantic sign scheme and finger spelling scheme.
A semantic sign scheme uses hand gestures, facial expressions, body parts, and actions to communicate meaning, tone, and sentiment.
A finger spelling scheme uses hand postures to represent alphabets in its corresponding language.
Automatic sign language transcription would allow better communication between a deaf community and hearing majority.
Sign language recognition has been subjects of various studies: visual-based approach\cite{NakjaiKatanyukul1, KishoreEtAl2016a, ChansriSrinonchat, PariwatSeresangtakul, Silanon} and sensory-glove-based approach\cite{InoueEtAl2015a, OzLeu2011a}.
A recent study\cite{NakjaiKatanyukul1}, investigating hand sign recognition for Thai Finger Spelling (TFS), has discussed issues and challenges in automatic transcription of TFS.
Although the discussion is based on TFS, some issues are general across languages or even general across domains beyond sign language recognition.
One of the key issues discussed in the study\cite{NakjaiKatanyukul1} is an issue of a non-sign or an invalid TFS sign, which may appear unintentionally during sign transition or from unaware hand postures. 

The appearance of non-signs may undermine the overall transcription performance.
Nakjai and Katanyukul\cite{NakjaiKatanyukul1} proposed
a light-weight computation approach 
to 
address the issue. 
Sign recognition is generally based on multi-class classification, whose output is represented in softmax coding.
That is, a softmax output capable of predicting one of $K$ classes is noted $\vec{y} = [y_1 y_2 \ldots y_K]^T$, whose coding bit $y_i \in [0,1], i = 1, \ldots, K$ and $\sum_{i=1}^K y_i = 1$.
A softmax output $\vec{y}$ represents predicted class $k$ when $y_k$ is the largest component: $k = \arg\max_i y_i$. 
%
Their approach is based on the assumption that the  ratio between the largest value of the coding bit and the rest shows the confidence of the model in its class prediction.
Softmax coding values have been normalized so that it can be associated to both probability interpretation and cross-entropy calculation.
Despite the benefits of normalization, 
they use the penultimate values instead of the softmax values 
for rationale that 
some information might have been lost during the softmax activation.
Penultimate values are inference values before going through softmax activation (i.e., $a_k$ in Eq.~\ref{eq: softmax y_k}).
Specifically, to indicate a non-sign posture, they proposed a confidence ratio, $cr = \frac{a}{b}$, where $a$ and $b$ are the largest and second largest penultimate values, respectively: $a = a_m$ and $b = a_n$ where $m = \arg\max_i {a_i}$ and $n = \arg\max_{i \neq m} {a_i}$.
Their confidence ratio has been reported to be effective in identifying a posture that is likely to get a wrong prediction. 
However, on their evaluating environment, they reported that the confidence ratio could hardly distinguish the cause of the wrong prediction whether it is a misclassified valid sign 
or it is a forced prediction on an invalid sign.
%
In addition, generally each penultimate output is a real number, $a_i \in \mathbb{R}$.
This nature poses a risk on confidence ratio formulation for when there is zero or a negative number, $cr$ can be misleading or its computation can even collapse (when denominator is zero).

Our study investigates development of an automatic hand sign recognition for Thai Finger Spelling (TFS), alternative formulations to confidence ratio, a non-sign issue and potential mitigations for a non-sign issue.
TFS, designated by the National Association of the Deaf in Thailand, has $25$ hand postures to represent $42$ Thai alphabets %
 using single-posture and multi-posture schemas\cite{NakjaiKatanyukul1}.
Single-posture schema directly associates a hand posture to a corresponding alphabet.
Multi-posture schema associates a series of $2$ to $3$ hand postures to a corresponding alphabet.
Based on probability interpretation of an inference output, Bayes theorem, and examining an internal structure of a commonly adopted inference model, 
various formulation alternatives to confidence ratio 
and 
a mitigation to a non-sign issue
are investigated (Section~\ref{sec: non-sign}).
Sections~\ref{sec: background},~\ref{sec: Experiments}, and~\ref{sec: conclusions} provide related background, methodologies and experimental results, and discussion and conclusions, respectively.

\section{Background}
\label{sec: background}

\paragraph{TFS hand sign recognition.} A recent visual-based state-of-the-art in TFS sign recognition\cite{NakjaiKatanyukul1} frames hand sign recognition as a pipeline of hand localization and sign classification problem.
They proposed an approach based on color scheme and contour area using Green's theorem for hand localization.
Then, an image region dominated by a hand is scaled to a pre-defined size (i.e., $64 \times 64$) and passed through a classifier, implemented with a convolution neural network.
The classifier predicts the most likely class out of the $25$ pre-defined classes, each corresponding to a valid TFS sign.

Most visual-based TFS sign recognition studies\cite{PariwatSeresangtakul, Silanon, NakjaiKatanyukul1} focus on a static image.
However, a practical system should anticipate video and streaming data, where unintended postures may be passed through the pipeline and cause confusion to the final transcription result.
Unintended postures can accidentally match valid signs.
This challenging case is worth further investigation and could be addressed through a language model.
However, even when the unintended postures do not match any of the valid signs,
a classifier is forced to predict one out of its pre-defined classes. 
No matter which class it predicts, the prediction is wrong. 
This could cause immediate confusion on its recognition result or this can undermine performance of its subsequence process in case of using this recognition as a part of a larger system.

\paragraph{Novelty, anomaly, outlier detections, and zero-shot learning.} 
A conventional classifier specifies a fixed number of classes it can predict and forced to predict.
This constraint allows it to be efficiently optimized to its classification task, but it has a drawback, which is more apparent when the assumption of all-inclusive classes is strongly violated. 
The concept of flagging out an instance belonging to a class that an inference machine has not seen at all in the training phase
is a common issue and a general concern beyond sign language recognition.
The issue has been extensively studied under various terms\footnote{
Definition of novelty, anomaly, outlier, and zero-shot 
may be slightly different.
Approaches may be various\cite{PimentelEtAl2014a, XianEtAl2017a},
but they are generally addressing a similar concern. 
}%
, e.g., 
novelty detection, anomaly detection, outlier detection, and zero-shot learning.

Pimentel et al.\cite{PimentelEtAl2014a} summarize a general direction in novelty detection.
That is, a detection method usually builds a model using training data containing no examples or very few examples of the novel classes.
Then, somehow depending on approaches, a novelty score $s$ is assigned to the data under question $\vec{x}$
and the final novelty judgement is decided by thresholding, i.e., the data $\vec{x}$ is judged a novelty (belonging to a new class) when $s(\vec{x}) > \tau$ for $\tau$ is a pre-defined threshold.

To obtain the novelty score, various approaches have been examined.
Pimentel et al.\cite{PimentelEtAl2014a} categorize novelty detection into $5$ approaches: probabilistic, distance-based, reconstruction-based, domain-based, and information-theoretic based techniques. 
Probabilistic approach 
relies on estimating probability density function (pdf) of the data.
A sample $\vec{x}$ is tested by thresholding the value of its pdf: $pdf(\vec{x}) < \tau$ indicates $\vec{x}$ being novelty.
Training data is used to estimate the pdf.
Although this approach has a strong theoretical support,
estimating pdf in practice requires a powerful generative model along with an efficient mechanism to train it.
Generative model at its fullest potential could provide greater inference capacities on data, such as expressive representation, reconstruction, speculation, generation, and structured prediction.
Its applicability is much beyond novelty detection.
However, high-dimension structured data renders this requirement very challenging.
A computationally traceable generative model is a subject of highly active research on its own right. 
Another related issue is to determine a sensible value for $\tau$, in which many studies\cite{Roberts2000a, CliftonEtAl2011a, BendaleBoult2015a} have resorted to extreme value theory (EVT)\cite{Pickands1975}.

Distance-based approach is presumably\cite{PimentelEtAl2014a} based on an assumption that 
data seen in a training process is tightly clustered
and data of new types locate far from their nearest neighbors in the data space.
Either a concept of nearest neighbors\cite{ZhangWang2006a} or of clustering\cite{KimEtAl2011a} is used.
Roughly speaking, a novelty score is defined either by a distance between a sample $\vec{x}$ and its nearest neighbors
or by a distance between $\vec{x}$ and its closest cluster centroids.
Distance is often measured with Euclidean or Mahalanobis distance.
The approach relies on mechanism to identify the nearest neighbors or the nearest clusters.
This usually is computationally intensive
and becomes a key factor attributed to its scalability issue in terms of data size and data dimensions.
Reconstruction-based approach involves building a re-constructive model, sometimes called ``auto-encoder,''
which learns to find a compact representation of input and reproduce it as an output.
Then, to test a sample, the sample is put through a reconstruction process
and a degree of dissimilarity between the sample and its reconstructed counterpart is used as a novelty score.
Hawkins et al.\cite{HawkinsEtAl2002a} uses a 3-hidden-layer ANN learned to reproduce its input.
Therefore, numbers of input and output nodes are equal.
The ANN is structured to have numbers of nodes in the hidden layers smaller than a number of input or output nodes in order to force ANN to learn compressed representation of the data.
Any sample that cannot be reconstructed well is taken for novelty, as this infers that its internal characteristics do not align with the compressed structure fine tuned to the training data. 
This approach may also resort to distance measurement for a degree of dissimilarity, 
but it does not require to search for the nearest neighbors.
Therefore, once a good auto-encoder is obtained, 
scalability is not as an issue as in
a distance-based approach.
Domain-based approach is related to building a boundary of the data domain in feature space.
Any sample is considered novelty if its location on feature space lies outside the boundary.
Sch\"olkopf et al.\cite{ScholkopfEtAl1999a} proposed one-class SVM for novelty detection.
Its key factors are to have SVM learn to build a boundary in feature space to adequately cover most training examples, 
while having a user-defined parameter to control a degree to allow some training samples to be outside the boundary. 
This compromising mechanism is a countermeasure to outliers in training data.
The last approach is information-theoretic.
It involves measurement of information content in the data.
It assumes that samples of novelty 
increase 
information content in the dataset significantly.
As their task is to remove outliers from data, 
He et al.\cite{HeEtAl2006a} uses a decrease in entropy of  a dataset after removal of the samples to indicate a degree that the samples are outliers.
The samples are heuristically searched.
Pimentel et al.\cite{PimentelEtAl2014a} noted that this approach often requires an information measure that is sensitive enough to pick up  the effect of novelty samples, especially when a number of these samples are small.

Based on this categorization\cite{PimentelEtAl2014a}, 
probabilistic approach is closest to the direction our work is taking.
However, unlike many early works, 
firstly, rather than requiring a dedicated model,
our proposed method builds upon a well-adopted classifier.
It can be used with an already-trained model without requirement for re-training.
Secondly, 
most works including a notable work of OpenMax\cite{BendaleBoult2015a}
determine a degree of novelty of a sample 
by
how unlikely it is to belong to any seen classes.
Another word, most previous works deduce probability of the sample being novelty by examining 
every probability of the sample being of the seen class, i.e., small values of $Pr[class = i|\vec{x}]$, for all $i = 1, \ldots, K$ are used to deduce the degree of novelty of $\vec{x}$.
Our work follows our interpretation of softmax output, i.e., $y_k \equiv Pr[class = k|s, \vec{x}]$, where $s$ represents a state of a valid sign (not novelty).
How likely sample $\vec{x}$ is a novelty then can be directly deduced.



\section{Prediction confidence and non-sign identification}
\label{sec: non-sign}



\paragraph{Confidence score.}
%
Since $y_k$ is associated with a probability of being in class $k$ and $y_k$ is generally obtained through a softmax mechanism (Eq.~\ref{eq: softmax y_k}), 
our study develops alternative formulations, as shown in Table~\ref{tbl: cs lc}.
Since $y_k$ has a direct probability interpretation, formulation $cs_1$ is straightforward.
Formulation $cs_2$ is associated to a logarithm of probability.
Formulation $cs_3$ is similar to confidence ratio%
, but it is an attempt to link an empirical utility to a theoretical rationale (probabilistic interpretation).
In addition, formulation $cs_3$ is preferably in term of computational cost and stability.
Formulation $cs_4$, a logit function, has a more direct interpretation of the starting assumption that the confidence is high when probability of the predicted class is much higher than the rest.
%


\begin{table}
\begin{center}
\caption{Formulations under investigation
for confidence score and lalent cognizance function. 
Softmax value $y_l = \frac{e^{a_l}}{\sum_{i=1}^K e^{a_i}}$, where $K$ is a number of predefined classes; 
$a_l$ is a penultimate value;
$k$ and $j$ are indices of the largest and the second largest components, respectively.
}
\label{tbl: cs lc}
\begin{tabular}{|c|c|c|c|c|c|}
\hline 
\multicolumn{2}{|c|}{Confidence}
& $cs_1 = y_k$ 
& $cs_2 = a_k$   
& $cs_3 = \log\left( \frac{y_k}{y_j} \right)$
& $cs_4 = \log\left( \frac{y_k}{1 - y_k} \right)$
\\
\multicolumn{2}{|c|}{score}
& & &  $= a_k - a_j$ &
\\ 
\hline
Latent 
& & & & &
\\
cognizance 
&
$\tilde{g}_0(a) = a$
&
$\tilde{g}_1(a) = e^a$
&
$\tilde{g}_2(a) = a^2$
&
$\tilde{g}_3(a) = a^3$
&
$\tilde{g}_4(a) = |a|$
\\ 
\hline 
\end{tabular} 
\end{center}
\end{table}



\paragraph{Latent cognizance.} 
%
Given the input image $\vec{x}$, the predicted sign in softmax coding $\vec{y} \in \mathbb{R}^K$ is derived through a softmax activation: 
for $k = 1, \ldots, K$,
\begin{eqnarray}
y_k = \frac{e^{a_k}}{\sum_{i=1}^K e^{a_i}}
\label{eq: softmax y_k},
\end{eqnarray}
where $a_k$ is the $k^{th}$ component of penultimate output, which has $K$ defined classes.
Each $y_k$ can be interpreted as a probability that the given image belongs to sign class $k$, 
or more precisely
a probability
that the given valid input belongs to class $k$.
That is, 
\begin{eqnarray}
y_k &\equiv& Pr[k|s, \vec{x}]
\label{eq: interpreted y_k}
\end{eqnarray}
where 
$k$ indicates one of the $K$ valid classes, 
$\vec{x}$ is the input under question,
and $s$ indicates that 
$\vec{x}$ is representing one of the valid classes (being a sign).
For conciseness, conditioning on $\vec{x}$ may be omitted, e.g., Eq.~\ref{eq: interpreted y_k} may be written as $y_k = Pr[k|s]$.
%
Noted that, this insight is distinct to
a common interpretation\cite{BendaleBoult2015a} that softmax coding bit $y_k$ of a well-learned inference model estimates probability of being in class $k$, i.e., $y_k = Pr[k|\vec{x}]$. 
This common notion does not emphasize
its conditioning on an inclusiveness of all pre-defined classes.


To identify a non-sign is another side of determining
the probability of being a sign: $Pr[\bar{s}|\vec{x}] = 1 - Pr[s|\vec{x}]$.
%
To deduce $Pr[s|\vec{x}]$, or concisely $Pr[s]$, 
consider Bayesian relation:
$Pr[k|s] = \frac{Pr[k, s]}{\sum_{i=1}^K Pr[i, s]}$
%
where $Pr[k, s]$ is a joint probability.
Given the Bayesian relation, inference mechanism (Eq.~\ref{eq: softmax y_k}), and our interpretation of $y_k$ (Eq.~\ref{eq: interpreted y_k}), 
the following relation is found:
\begin{eqnarray}
\frac{e^{a_k}}{\sum_{i=1}^K e^{a_i}} &=& \frac{Pr[k, s]}{\sum_{i=1}^K Pr[i, s]}
\label{eq: fundamental cognizance}.
\end{eqnarray}

It is noticeable that term $e^{a_k}$ is in the same structure as joint probability $Pr[k, s]$ is.
Here, we draw the assumption that penultimate value $a_k$ relates to joint probability $Pr[k, s]$ through an unknown function $u: a_k(\vec{x}) \mapsto Pr[k, s|\vec{x}]$.
Theoretically, this unknown function is difficult to exactly characterize.
In practice, even without exact characteristics of this mapping, a good approximate is enough to accomplish a task of identifying a non-sign.
Supposed there exists an approximate mapping $g$, i.e., $g(a_k) \approx Pr[k, s]$,
therefore given $g(a_i)$'s, a non-sign can be identified by
$Pr[s|\vec{x}] = \sum_i Pr[i, s|\vec{x}] \approx \sum_i g(a_i(\vec{x}))$.
Further refining, to lessen burden on enforcing proper probability properties on $g$,
define a ``latent cognizance'' function $\tilde{g}$ such that $\tilde{g}(a_i(\vec{x})) \propto g(a_i(\vec{x}))$.
Consequently, define primary and secondary latent cognizances as the following relations, respectively:
\begin{eqnarray}
\tilde{g}\left(a_i(\vec{x})\right) & \propto & Pr[i, s|\vec{x}]
\label{eq: cognizance 1}, \\
\sum_i \tilde{g}\left(a_i(\vec{x})\right) & \propto & Pr[s|\vec{x}]
\label{eq: cognizance 2}.
\end{eqnarray}

Various formulations (Table~\ref{tbl: cs lc}) are investigated for an effective latent cognizance function.
Identity $\tilde{g}_0$ is chosen for its simplicity.
Exponential $\tilde{g}_1$ is chosen for its immediate reflection on Eq.~\ref{eq: fundamental cognizance}.
It should be noted that a study on a whole family of $\tilde{g} = m \cdot e^a$, where $m$ is a constant, is worth further investigation.
Other formulations are 
intuitively
included on an exploratory purpose.

\section{Experiments}
\label{sec: Experiments}

\begin{figure}
\begin{center}
\includegraphics[width=\textwidth]{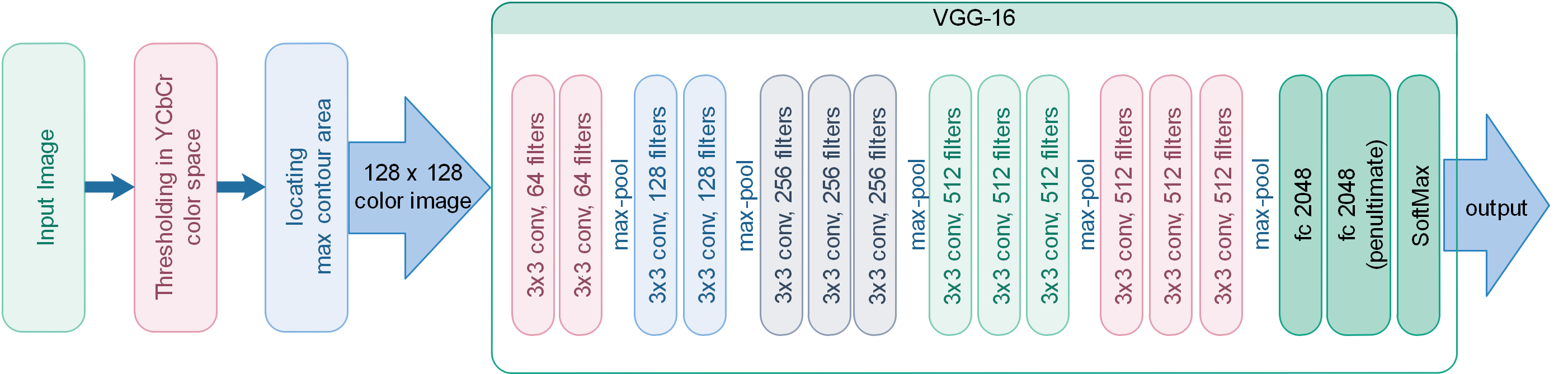}
\caption{Processing pipeline of our TFS sign recognition.} 
\label{fig: TFS Sign Recognition}
\end{center}
\end{figure}

Various formulations of confidence score and choices of latent cognizance are evaluated on
TFS sign recognition system.
Our TFS sign recognition follows the current state-of-the-art in visual TFS sign recognition\cite{NakjaiKatanyukul1}
with a modification of convolution neural network (CNN) configuration and its input resolution.
Instead of a $64 \times 64$ gray-scale image,
our work uses a $128 \times 128$ color image as an input for CNN.  
Our CNN configuration uses a VGG-16\cite{VGG-16} 
with the $2$ fully-connected layers each having $2048$ nodes, instead of $3$ fully-connected layers in the original VGG-16.
%
Fig.~\ref{fig: TFS Sign Recognition} illustrates a processing pipeline of our TFS sign recognition.

\paragraph{Sign data.}
The main dataset contains images of 25 valid TFS sign postures.
Each valid TFS sign data is collected from 12 signers and posed $5$ times by each signer.
That results in a total number of $1500$ images ($5$ times $\times 25$ postures $\times 12$ signers), which are augmented to $15000$ images.
All augmented images are visually inspected for human readability and semantic integrity.
Every image is a color image with a resolution of approximately $800 \times 600$ pixels.

\paragraph{Experimentation.}
The data is separated by signer into training set ($11250$ images from $9$ signers making up as $75\%$) and test set (the $3750$ images from the $3$ signers, making up $25\%$).
The experiments are conducted for $10$ repetitions in a $10$-fold manner.
Specifically, each repetition separates data differently, e.g., the $1^{st}$ fold uses data from signers 1, 2, and 3 for test and uses the rest for training;
the $2^{nd}$ fold uses data from signers 2, 3, and 4 for test; and so on till the last fold using data from signers 10, 1, and 2 for test.

The mean Average Precision (mAP), commonly used in object detection\cite{Yolo1}, is 
a key performance measurement for evaluation of our TFS sign recognition.
Area under curve (AUC)
and receiver operating characteristic (ROC) are used to evaluate effectiveness of various formulations for confidence score and latent cognizance.
AUC is often referred to  an estimate area under Precision-Recall curve,
while ROC is usually referred to an estimate area under Detection-Rate--False-Alarm-Rate curve.
However, generally both areas are equivalent.
We use them to differentiate 
the purpose of our evaluation rather than taking them as different performance metrics.
AUC is used for identification of samples not to be correctly predicted\footnote{
Positive is defined to be a sample of either a non-sign or an incorrect prediction.
}.
It is more direct to measure a quality 
of a replacement for the confidence ratio.
ROC
is used for identifying non-sign samples%
\footnote{
Positive is defined to be a non-sign.
}.
It is more direct to the very issue of non-sign postures.

\paragraph{Non-sign data.}
In addition to the sign dataset, a non-sign dataset containing mixed-posing invalid TFS sign postures is used to evaluate non-sign identification methods.
The invalid TFS posture data is collected from a signer and augmented to $1122$ images.
All has been visually inspected that they all are readable and do not accidentally match to any of the $25$ valid signs.

\begin{figure}
\includegraphics[width=\textwidth]{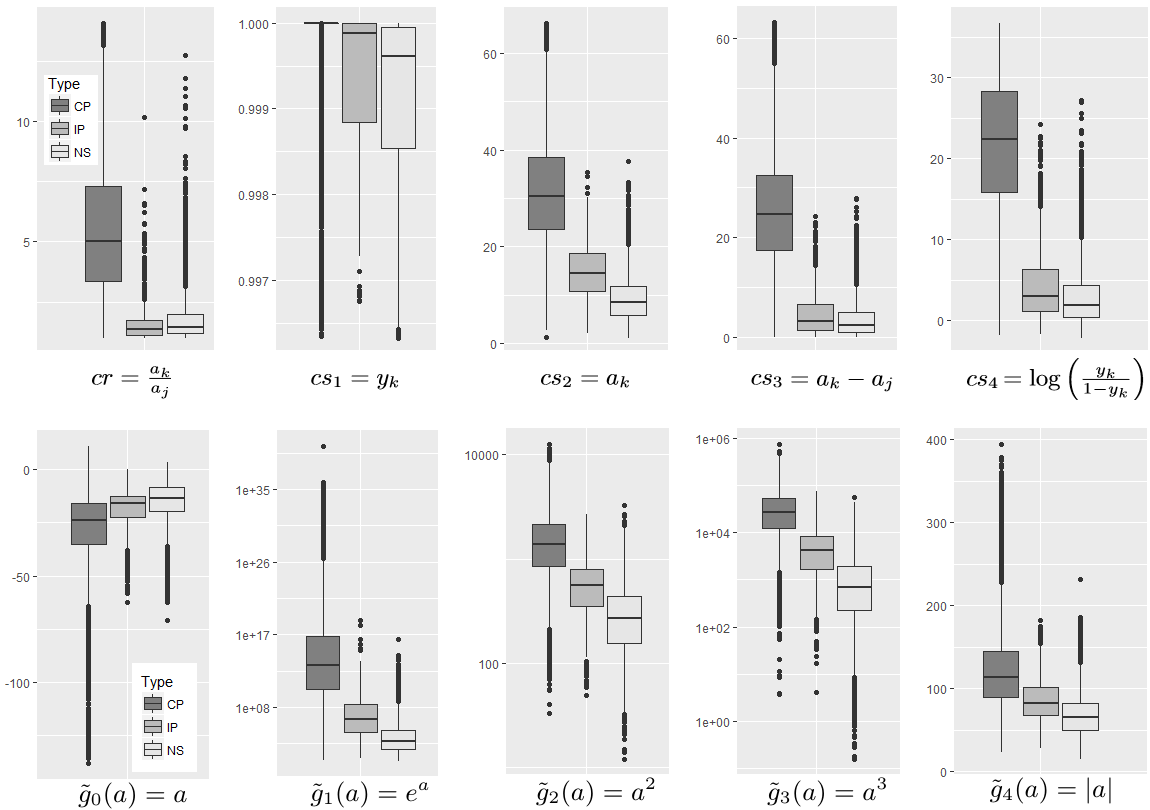}
\caption{Upper:
Boxplots of confidence ratio and candidates for confidence score.
Y-axis shows treatment values in linear scale.
Lower:
Boxplots of $5$ candidates for latent cognizance function. 
Y-axis shows $\sum_i \tilde{g}(a_i)$ values ($\tilde{g}_0$ and $\tilde{g}_4$ in linear scale; the rest in log scale).
Boxplots are shown in $3$ groups: 
\texttt{CP} for correctly classified samples;
\texttt{IP} for misclassified samples;
\texttt{NS} for non-sign samples.
}
\label{fig: boxplots}
\end{figure}

\paragraph{Results.}
Table~\ref{tbl: perf sign recog} shows TFS recognition performance of the previous studies and our work.
The high performing mAP ($97.59\%$) indicates that the model is well-learned. 
All data are shown to be non-normal distributed, based on Lilliefors test at $0.05$ level.
Wilcoxon rank-sum test is conducted on each treatment for comparing (1) difference between \verb|CP| and \verb|IP|,
(2) difference between \verb|CP| and \verb|NS|,
and (3) difference between \verb|IP| and \verb|NS|.
The notations \verb|CP|, \verb|IP|, and \verb|NS| represent samples being correctly predicted, being misclassified, and being a non-sign, respectively.
At $0.01$ level, Wilcoxon rank-sum test confirms all $3$ differences in all treatments (including confidence ratio, $4$ formulations for confidence score, and $5$ latent cognizance functions).
Figure~\ref{fig: boxplots} shows boxplots of all treatments.
Although the significance tests confirm that 
the $3$ groups can be distinguishable using any of the treatments, the boxplots show a wide range of degrees of distinguishment, e.g., cognizance $\tilde{g} = a^3$ ($4^{th}$ plot in Fig.~\ref{fig: boxplots}) seems to be easier than others on thresholding the $3$ cases.
To measure a degree of effectiveness,
Tables~\ref{tbl: perf conf score}
~and~\ref{tbl: perf latent cog} provide
AUC and ROC for various methods under investigation.
Noted that, since treatment $\tilde{g}_0$ gives results in a different manner than others: a higher value associates to a non-sign (c.f. a lower value in others),
the evaluation logic is adjusted accordingly.

\begin{figure}
\begin{tabular}{cc}
\includegraphics[width=0.5\textwidth]{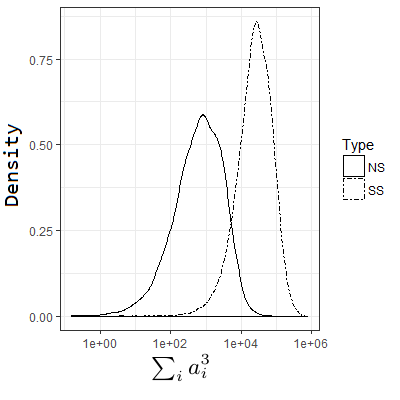}
&
\includegraphics[width=0.5\textwidth]{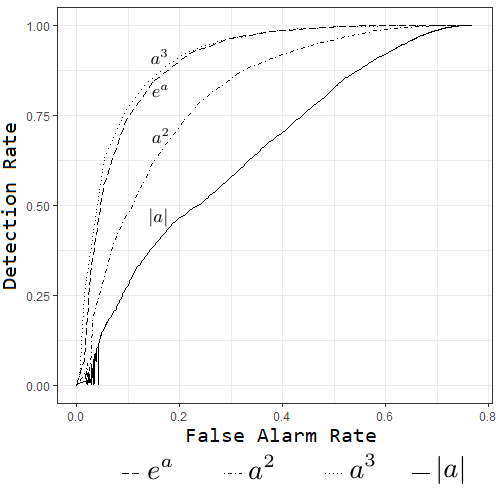}
\end{tabular} 
\caption{(a) Illustration of smoothed estimated densities of sign (denoted \texttt{SS}) and non-sign (denoted \texttt{NS}) data over $\sum_i a_i^3$.
(b) Detection rate versus false alarm rate curves of the $4$ strongest candidates.
} 
\label{fig: latent cognizance}
\end{figure}

On finding an alternative to confidence ratio,
maximal penultimate output $a_k$ seems to be a good replacement
such that it provides the largest AUC ($0.934$)
and 
it is simple to obtain (no extra computation, thus no risk of computational instability).
On addressing a non-sign issue, latent cognizance with cubic function $\tilde{g}(a) = a^3$ gives the best ROC ($0.929$).
Its smoothed estimate densities%
\footnote{
A normalized Gaussian-smoothing version of histogram
produced through smoothed density estimates of \texttt{ggplot2} 
(\url{http://ggplot2.tidyverse.org})
with default parameters.
} 
of non-sign samples (\verb|NS|) and sign samples (combining \verb|CP| and \verb|IP|) are
shown on the left of Fig.~\ref{fig: latent cognizance}.
Plots of detection rate and false alarm rate of the $4$ strongest candidates are shown on the right of Fig.~\ref{fig: latent cognizance}.

\begin{table}
{\scriptsize
\begin{center}
\caption{Performance of visual-based TFS sign recognition. 
}
\label{tbl: perf sign recog}
\begin{tabular}{|l|c|r|l|c|}
\hline
Method & TFS      & Data Size & Key     & Performance
\\
       & Coverage & (\# images) & factors &
\\
\hline
\hline
Chansri and Srinonchat\cite{ChansriSrinonchat}
& $16$ signs & $320$ & Kinect 3D camera, HOG and ANN & $83.33\%$ \\
\hline 
Pariwat and Seresangtakul\cite{PariwatSeresangtakul}
& $15$ signs & $375$ & SVM & $91.20\%$ \\
\hline
Silanon\cite{Silanon} & $21$ signs & $2100$ & HOG and ANN & $78.00\%$ \\
\hline
Nakjai and Katanyukul\cite{NakjaiKatanyukul1}
& $25$ signs & $1375$ & Hand Extraction and CNN & $91.26\%$ \\
\hline
Our work
& $25$ signs & $15000$ & Hand Extraction and VGG-16 &  97.59\% \\
\hline
\end{tabular} 
\end{center}
}%
\end{table}

\begin{table}
\begin{center}
\caption{Evaluation of confidence score formulations. 
}
\label{tbl: perf conf score}
\begin{tabular}{|l|c|c|c|c|c|}
\hline 
&
$cr = \frac{a_k}{a_j}$
&
$cs_1 = y_k$ & $cs_2 = a_k$ 
& $cs_3 = a_k - a_j$
& $cs_4 = \log\left( \frac{y_k}{1 - y_k} \right)$
\\ 
\hline
AUC & $0.814$ & $0.919$ & $0.934$ 
& $0.900$ & $0.919$
\\
\hline
ROC & $0.740$ & $0.879$ & $0.921$ 
& $0.847$ & $0.879$
\\
\hline
\end{tabular} 
\end{center}
\end{table}

\begin{table}
\begin{center}
\caption{Evaluation of various $\tilde{g}$ formulations on $\sum_i \tilde{g}(a_i) \propto Pr[s]$. 
}
\label{tbl: perf latent cog}
\begin{tabular}{|l|c|c|c|c|c|}
\hline
&
Identity
&
Exponential
&
Quadratic
&
Cubic
&
Absolute
\\
&
$\tilde{g}_0(a) = a$
&
$\tilde{g}_1(a) = e^a$
&
$\tilde{g}_2(a) = a^2$
&
$\tilde{g}_3(a) = a^3$
&
$\tilde{g}_4(a) = |a|$
\\ 
\hline
AUC & $0.437$ & $0.930$ & $0.855$ & $0.934$ & $0.737$ 
\\
\hline
ROC
& $0.419$ & $0.920$ & $0.845$ & $0.929$ & $0.726$ \\
\hline
\end{tabular} 
\end{center}
\end{table}

\section{Discussion and Conclusions}
\label{sec: conclusions}

The cubic function $\tilde{g}(a) = a^3$ has shown to be 
the best cognizance function among other candidates, including exponential 
$\tilde{g}(a) = e^a$.
In addition, the cubic cognizance has ROC par to 
the max-penultimate confidence score.
On the other hand, 
the max-penultimate confidence score
also provide a competitive ROC and could be used to identify non-sign samples as well.
Noted that OpenMax---a notable work in novelty detection---uses penultimate output as one of its crucial parts to identify novelties.
Our finding could contribute to the development of OpenMax. 
A study of using cubic cognizance in OpenMax system may be another promising direction, since it is shown to be more effective than penultimate output.
Another point worth noting is that 
the previous work\cite{NakjaiKatanyukul1}
evaluated confidence score on identifying non-signs
and could not confirm its effectiveness
with the significance tests.
Their results agree with our early experiments when using a lower resolution image, a smaller CNN structure,
and training and testing on smaller datasets.
In our early experiment, only a few of the treatments could be confirmed for non-sign identification.
Those that were confirmed are consistent with ROC presented here.
This observation implies a strong relation between state of the inference model and non-sign-identification effectiveness. 
This relation and its details should be systematically investigated.
Regarding applications of the techniques,
thresholding can be used
and
a proper value for threshold is needed to be determined.
This can be simply achieved through tracing the Detection-Rate--False-Alarm-Rate curve with the corresponding threshold values.
Alternatively, the proper threshold can be determined based on Extreme Value Theory, like many previous studies\cite{Roberts2000a, CliftonEtAl2011a, BendaleBoult2015a}.
Another interesting research direction is to find a similar solution for other inference families.
Our techniques target a softmax-based classifier,
which is well-adopted especially in artificial neural network.
However, Support Vector Machine (SVM), another well-adopted classifier, is built on a different paradigm.
Application of either confidence score or latent cognizance to SVM might not work or might be totally irrelevant.
Investigation into the issue on other inference paradigms could provide a unified insight of the underlying inference mechanism and it could be beneficial beyond addressing the novelty issue.
Regarding starting assumptions, high ROC values of exponential and cubic cognizances support our new interpretation 
and its following assumptions.
However, the penultimate output, according to our new interpretation, has relation $a_k(\vec{x}) = \log(Pr[k|s,\vec{x}]) + C$,
where $C = - \log \sum_i a_i(\vec{x})$.  
This relation only partially agrees with the experimental results.
High value of AUC agrees with $\log(Pr[k|s,\vec{x}])$
that a class is confidently classified,
but $Pr[k|s,\vec{x}]$ alone is not enough to determine a non-sign, which needs $Pr[\bar{s}|\vec{x}]$.
This may imply that our research is going on a right direction, but it still needs more investigation to complete the picture.

In brief, our study investigates (1) alternatives to confidence ratio
and (2) methods to identify a non-sign.
The max-penultimate output is shown to be a good replacement for confidence ratio 
in terms of detection performance and simplicity.
Its large value associates to a sample likely to be correctly classified and vice versa.
The cognizance $\sum_i a_i^3$ is shown to be a good indicator for a non-sign
such that $\sum_i a_i^3(\vec{x}) \propto Pr[s|\vec{x}]$, or low value of $\sum_i a_i^3(\vec{x})$ associates to a non-sign sample.
%
To warp up our article, our findings give an insight into a softmax-based inference machine and provide a tool to measure a degree of confidence in the prediction result
as well as a tool to identify a novelty or an anomaly.
The implications may go beyond our current scope of TFS hand-sign recognition and contribute to novelty detection, open-set recognition, and other similar concepts.
Latent cognizance is very satisfactory for its simplicity and effectiveness in identifying non-signs.
These would help improve an overall quality of the translation, which in turn hopefully leads to a better understanding among people of different physical backgrounds.



%
%
%
\bibliographystyle{splncs04}
\bibliography{mybibliography}

%
%
\end{document}